\def\year{2020}\relax
\documentclass[letterpaper]{article} 
\usepackage{aaai20}  
\usepackage{times}  
\usepackage{helvet} 
\usepackage{courier}  
\usepackage[hyphens]{url}  
\usepackage{graphicx} 
\usepackage{graphicx}  
\usepackage{booktabs}
\title{Non-local U-Nets for Biomedical Image Segmentation}
\author{
Zhengyang Wang,\textsuperscript{\rm 1}
Na Zou,\textsuperscript{\rm 1}
Dinggang Shen,\textsuperscript{\rm 2}
Shuiwang Ji\textsuperscript{\rm 1}\\
\textsuperscript{\rm 1}Texas A\&M University,
\textsuperscript{\rm 2}University of North Carolina at Chapel Hill\\
zhengyang.wang@tamu.edu, nzou1@tamu.edu, dgshen@med.unc.edu, sji@tamu.edu
}
 
\begin{document}

\maketitle

\begin{abstract}
Deep learning has shown its great promise in various biomedical image segmentation tasks. Existing models are typically based on U-Net and rely on an encoder-decoder architecture with stacked local operators to aggregate long-range information gradually. However, only using the local operators limits the efficiency and effectiveness. In this work, we propose the non-local U-Nets, which are equipped with flexible global aggregation blocks, for biomedical image segmentation. These blocks can be inserted into U-Net as size-preserving processes, as well as down-sampling and up-sampling layers. We perform thorough experiments on the 3D multimodality isointense infant brain MR image segmentation task to evaluate the non-local U-Nets. Results show that our proposed models achieve top performances with fewer parameters and faster computation.
\end{abstract}

\section{Introduction}

In recent years, deep learning methods, such as fully convolutional networks~(FCN)~\cite{long2015fully}, U-Net~\cite{ronneberger2015u}, Deeplab~\cite{chen2018deeplab,wang2018smoothed}, and RefineNet~\cite{lin2017refinenet}, have continuously set performance records on image segmentation tasks. In particular, U-Net has served as the backbone network for biomedical image segmentation. Basically, U-Net is composed of a down-sampling encoder and an up-sampling decoder, along with skip connections between them. It incorporates both local and global contextual information through the encoding-decoding process.

Many variants of U-Net have been developed and they achieved improved performance on biomedical image segmentation tasks. For example, residual deconvolutional network~\cite{fakhry2017residual} and residual symmetric U-Net~\cite{lee2017superhuman} addressed the 2D electron microscopy image segmentation task by building a U-Net based network with additional short-range residual connections~\cite{he2016deep}. In addition, U-Net was extended from 2D to 3D cases for volumetric biomedical images, leading to models like 3D U-Net~\cite{cciccek20163d}, V-Net~\cite{milletari2016v}, and convolution-concatenate 3D-FCN~(CC-3D-FCN)~\cite{nie20183}.

Despite the success of these studies, we conduct an in-depth study of U-Net based models and observe two limitations shared by them. First, the encoder usually stacks size-preserving convolutional layers, interlaced with down-sampling operators, to gradually reduce the spatial sizes of feature maps. Both convolutions and down-sampling operators are typically local operators, which apply small kernels to scan inputs and extract local information. Stacking them in a cascade way results in large effective kernels and is able to aggregate long-range information. As the biomedical image segmentation usually benefits from a wide range of contextual information, most prior models have a deep encoder, \emph{i.e.,} an encoder with many stacked local operators. It hurts the efficiency of these models by introducing a considerably large amount of training parameters, especially when more down-sampling operators are employed, since the number of feature maps usually gets doubled after each down-sampling operation. In addition, more down-sampling operators cause the loss of more spatial information during encoding, which is crucial for biomedical image segmentation. Second, the decoder is built in a similar way to the encoder, by replacing down-sampling operators with up-sampling operators. Popular up-sampling operators, like deconvolutions and unpooling layers, are local operators as well~\cite{gao2019pixel}. However, the up-sampling process involves the recovery of spatial information, which is hard without taking global information into consideration. To conclude, it will improve both the effectiveness and efficiency of U-Net based models to develop a new operator capable of performing non-local information aggregation. As U-Net has size-preserving processes, as well as down-sampling and up-sampling layers, the new operator is supposed to be flexible to fit these cases.

In this work, we address the two limitations and propose the non-local U-Nets for biomedical image segmentation. To address the first limitation above, we propose a global aggregation block based on the self-attention operator~\cite{vaswani2017attention,wang2018non,yuan2019learn}, which is able to aggregate global information without a deep encoder. This block is further extended to an up-sampling global aggregation block, which can alleviate the second problem. To the best of our knowledge, we are the first to make this extension. We explore the applications of these flexible global aggregation blocks in U-Net on the 3D multimodality isointense infant brain magnetic resonance~(MR) image segmentation task. Experimental results show that our proposed non-local U-Nets are able to achieve the top performance with fewer parameters and faster computation.

\section{Non-local U-Net}

In this section, we introduce our proposed non-local U-Nets. We first illustrate the specific U-Net framework used by our models. Based on the framework, our models are composed of different size-preserving, down-sampling and up-sampling blocks. We describe each block and propose our global aggregation blocks to build the non-local U-Nets.

\subsection{U-Net Framework}

\begin{figure}[!t]
	\centering
	\includegraphics[width=0.95\columnwidth]{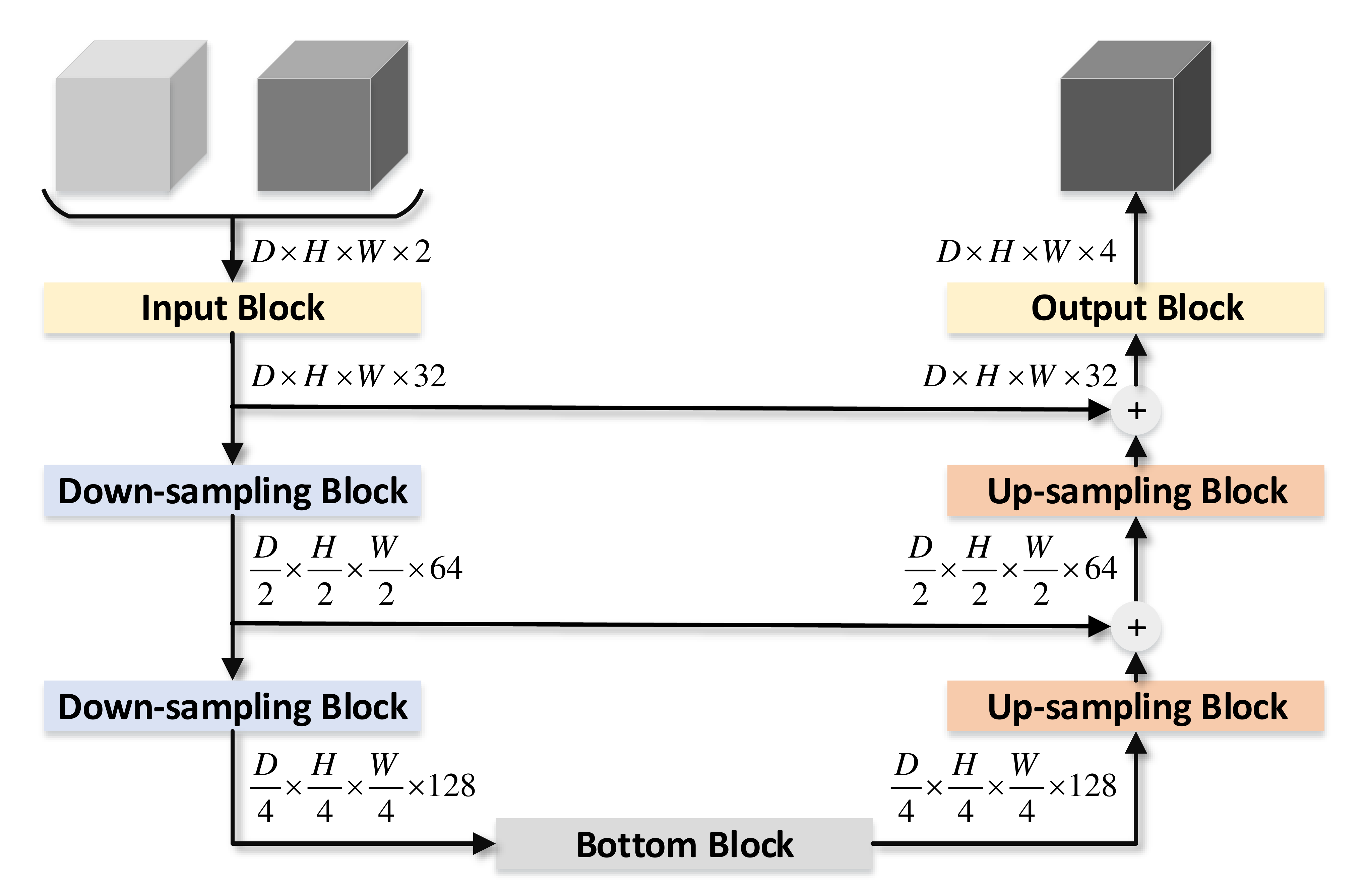}
	\caption{An illustration of the U-Net framework employed by our proposed non-local U-Nets.
		In this example, the inputs have 2 channels and the segmentation task has 4 classes.}
	\label{fig:unet}
\end{figure}

We describe the non-local U-Nets in 3D cases. Lower or higher dimensional cases can be easily derived.
An illustration of the basic U-Net framework is given in
Fig.~\ref{fig:unet}. The input first goes through an encoding input block,
which extracts low-level features. Two down-sampling blocks
are used to reduce the spatial sizes and obtain high-level
features. Note that the number of channels is doubled after each
down-sampling block. A bottom block then aggregates global
information and produces the output of the encoder. Correspondingly,
the decoder uses two up-sampling blocks to recover the spatial sizes
for the segmentation output. The number of feature maps is halved
after an up-sampling operation.

To assist the decoding process, skip connections copy feature maps
from the encoder to the decoder. Differently, in the non-local U-Nets, the copied
feature maps are combined with decoding feature maps through
summation, instead of concatenation used in U-Net~\cite{ronneberger2015u,yuan2018computational}. The
intuitive way to combine features from the encoder and the decoder
is concatenation, providing two sources of inputs to the up-sampling
operation. Using summation instead has two advantages~\cite{lin2017feature}. First,
summation does not increase the number of feature maps, thus
reducing the number of trainable parameters in the following layer.
Second, skip connections with summation can be considered as
long-range residual connections, which are known to be capable of
facilitating the training of models.

Given the output of the decoder, the output block produces the segmentation probability
map. Specifically, for each voxel, the probabilities that it belongs
to each segmentation class are provided, respectively. The final
segmentation map can be obtained through a single \textit{argmax}
operation on this probability map. The details of each block are
introduced in following sections.

\begin{figure*}[!t]
	\centering
	\includegraphics[width=1.6\columnwidth]{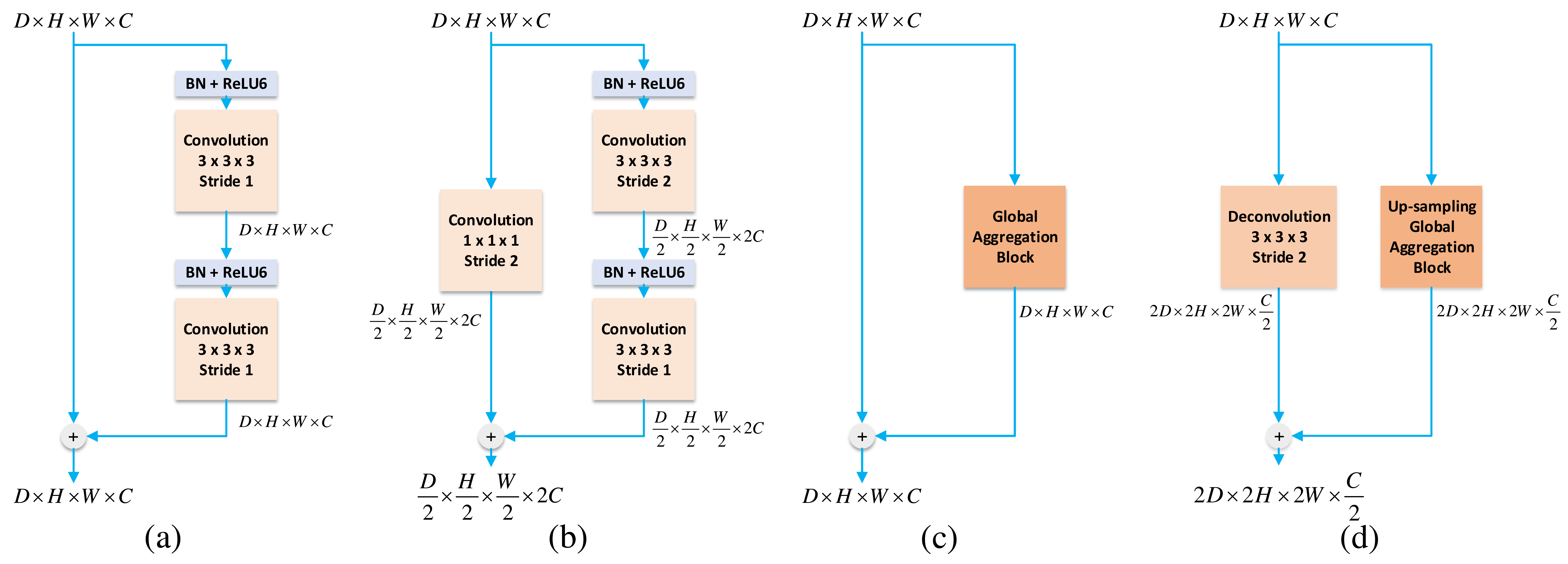}
	\caption{An illustration of the residual blocks employed by our proposed non-local U-Nets.
		Details are provided in Section ``Residual Block''.}
	\label{fig:residual}
\end{figure*}

\subsection{Residual Blocks}

Residual connections have been shown to facilitate the training of
deep learning models and achieve better
performance~\cite{he2016deep}. Note that skip connections with
summation in our U-Net framework are equivalent to long-range
residual connections. To further improve U-Net, the studies
in~\cite{lee2017superhuman,lin2017refinenet,fakhry2017residual} proposed to add short-range residual
connections as well. However, those studies did not apply residual connections for
down-sampling and up-sampling blocks. Down-sampling block with
residual connections has been explored in ResNet~\cite{he2016deep}.
We explore the idea for up-sampling blocks based on our proposed
up-sampling global aggregation block, as discussed in next section.

In our proposed model, four different residual blocks are used to form a
fully residual network, as shown in
Fig.~\ref{fig:residual}. Notably, all of them apply the pre-activation
pattern~\cite{he2016identity}. Fig.~\ref{fig:residual}(a) shows a regular
residual block with two consecutive convolutional layers. Here, batch
normalization~\cite{ioffe2015batch} with the
ReLU6 activation function is used before each convolutional
layer. This block is used as the input block in our framework. The output
block is constructed by this block followed by a $1 \times 1 \times 1$
convolution with a stride of 1. Moreover, after the summation of skip
connections, we insert one such block. Fig.~\ref{fig:residual}(b) is a
down-sampling residual block. A $1 \times 1 \times 1$ convolution with a
stride of 2 is used to replace the identity residual connection, in order to
adjust the spatial sizes of feature maps accordingly. We employ this block as
the down-sampling blocks. Fig.~\ref{fig:residual}(c) illustrates our bottom
block. Basically, a residual connection is applied on the proposed global
aggregation block. The up-sampling residual block is provided in
Fig.~\ref{fig:residual}(d). Similar to the down-sampling block in
Fig.~\ref{fig:residual}(b), the identity residual connection is replaced by
a $3 \times 3 \times 3$ deconvolution with a stride of 2 and the other branch
is the up-sampling global aggregation block. Our model uses this block as the
up-sampling blocks.

\subsection{Global Aggregation Block}

To achieve global information fusion through a block, each position
of the output feature maps should depend on all positions of the
input feature maps. Such an operation is opposite to local
operations like convolutions and deconvolutions, where each output
location has a local receptive field on the input. In fact, a
fully-connected layer has this global property. However, it is prone
to over-fitting and does not work well in practice. We note that the self-attention block used in the
Transformer~\cite{vaswani2017attention} computes outputs at one position by attending to every
position of the input. Later, the study in~\cite{wang2018non}
proposed non-local neural networks for video classification, which
employed a similar block. While both studies applied self-attention
blocks with the aim of capturing long-term dependencies in
sequences, we point out that global information of image feature
maps can be aggregated through self-attention blocks.

Based on this insight, we propose the global aggregation block,
which is able to fuse global information from feature maps of any
size. We further generalize it to handle down-sampling and
up-sampling, making it a block that can be used anywhere in deep
learning models.

\begin{figure*}[!t]
	\centering
	\includegraphics[width=1.6\columnwidth]{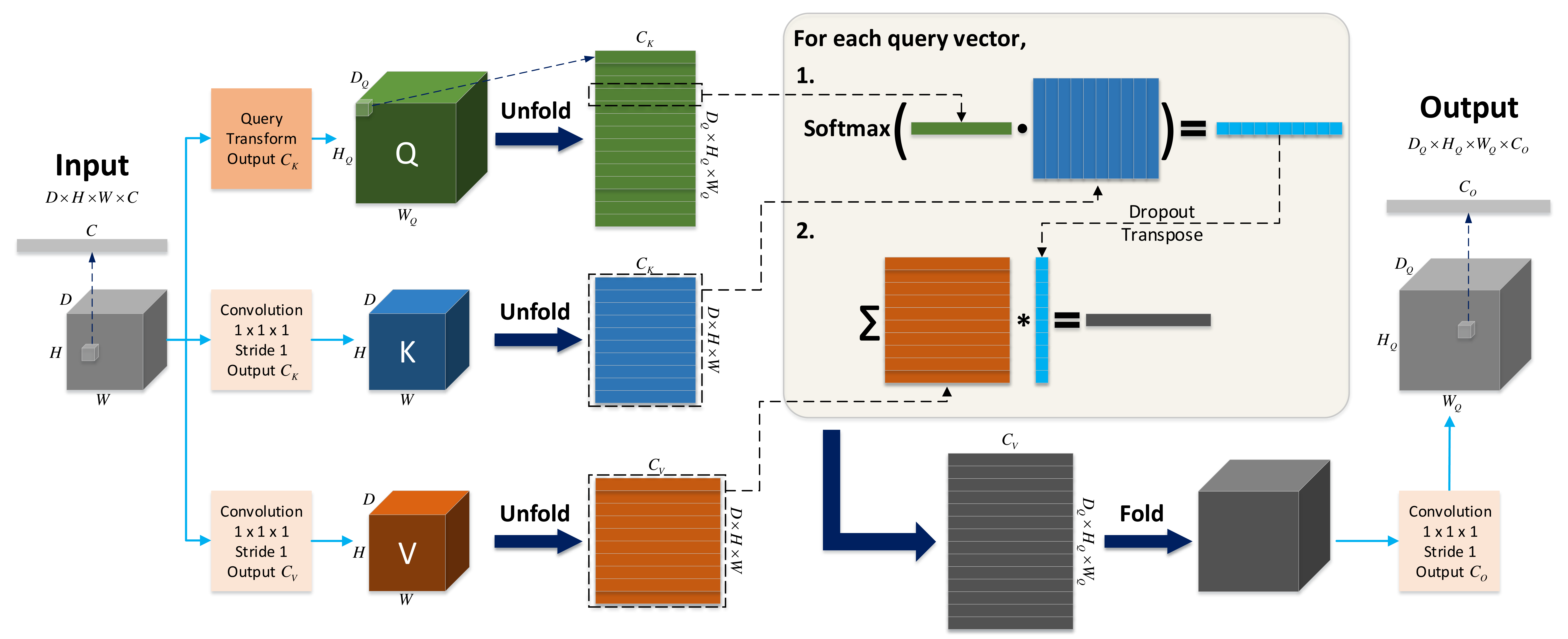}
	\caption{An illustration of our proposed global aggregation block. Note that
		the spatial size of the output is determined by that of the query~(Q) matrix.}
	\label{fig:global}
\end{figure*}

Let $X$ represent the input to the global
aggregation block and $Y$ represent the output. For simplicity, we
use $Conv\_1_{N}$ to denote a $1 \times 1 \times 1$ convolution with
a stride of 1 and $N$ output channels. Note that $Conv\_1_{N}$ does
not change the spatial size. The first step of the proposed block is
to generate the query~($Q$), key~($K$) and value~($V$)
matrices~\cite{vaswani2017attention},
given by
\begin{eqnarray}\label{eqn:step1}
Q&=&Unfold(QueryTransform_{C_K}(X)), \nonumber \\
K&=&Unfold(Conv\_1_{C_K}(X)), \nonumber \\
V&=&Unfold(Conv\_1_{C_V}(X)),
\end{eqnarray}
where $Unfold(\cdot)$ unfolds a $D \times H \times W \times C$
tensor into a $(D \times H \times W) \times C$ matrix,
$QueryTransform_{C_K}(\cdot)$ can be any operation that produces
$C_K$ feature maps, and $C_K, C_V$ are hyper-parameters representing
the dimensions of the keys and values. Suppose the size of $X$ is $D
\times H \times W \times C$. Then the dimensions of $K$ and $V$ are
$(D \times H \times W) \times C_K$ and $(D \times H \times W) \times
C_V$, respectively. The dimension of $Q$, however, is $(D_Q \times
H_Q \times W_Q) \times C_K$, where $D_Q, H_Q, W_Q$ depend on
$QueryTransform(\cdot)$. The left part of Fig.~\ref{fig:global}
illustrates this step. Here, a $D \times H \times W \times C$ tensor
is represented by a $D \times H \times W$ cube, whose voxels
correspond to $C$-dimensional vectors.

Each row of the $Q$, $K$ and $V$ matrices denotes a query vector, a
key vector and a value vector, respectively. Note that the query
vector has the same dimension as the key vector. Meanwhile, the
number of key vectors is the same as that of value vectors, which
indicates a one-to-one correspondence. In the second step, the attention
mechanism is applied on $Q$, $K$ and $V$~\cite{vaswani2017attention}, defined as
\begin{eqnarray}
A&=&Softmax(\frac{QK^T}{\sqrt{C_K}}), \nonumber \\
O&=&AV,
\end{eqnarray}
where the dimension of the attention weight matrix $A$ is $(D_Q
\times H_Q \times W_Q) \times (D \times H \times W)$ and the
dimension of the output matrix $O$ is $(D_Q \times H_Q \times W_Q)
\times C_V$. To see how it works, we take one query vector from $Q$
as an example. In the attention mechanism, the query vector
interacts with all key vectors, where the dot-product between the
query vector and one key vector produces a scalar weight for the
corresponding value vector. The output of the query vector is a
weighted sum of all value vectors, where the weights are normalized
through $Softmax$. This process is repeated for all query vectors
and generates $(D_Q \times H_Q \times W_Q)$ $C_V$-dimensional
vectors. This step is illustrated in the box of
Fig.~\ref{fig:global}. Note that
Dropout~\cite{srivastava2014dropout} can be applied on $A$ to avoid
over-fitting. As shown in Fig.~\ref{fig:global}, the final step of
the block computes $Y$ by
\begin{equation}
Y=Conv\_1_{C_O}(Fold(O)),
\end{equation}
where $Fold(\cdot)$ is the reverse operation of $Unfold(\cdot)$ and $C_O$
is a hyper-parameter representing the dimension of the outputs.
As a result, the size of $Y$ is $D_Q \times H_Q \times W_Q \times C_O$.

In particular, it is worth noting that the spatial size of $Y$ is determined by
that of the Q matrix, \emph{i.e.}, by the
$QueryTransform_{C_K}(\cdot)$ function in~(\ref{eqn:step1}). Therefore, with appropriate
$QueryTransform_{C_K}(\cdot)$ functions, the global aggregation
block can be flexibly used for size-preserving, down-sampling and
up-sampling processes. In our proposed non-local U-Nets, we set $C_K=C_V=C_O$ and explore two different $QueryTransform_{C_K}(\cdot)$ functions.
For the global aggregation block in Fig.~\ref{fig:residual}(c),
$QueryTransform_{C_K}(\cdot)$ is $Conv\_1_{C_K}$. For the up-sampling global
aggregation block in Fig.~\ref{fig:residual}(d),
$QueryTransform_{C_K}(\cdot)$ is a $3 \times 3 \times 3$
deconvolution with a stride of 2. The use of this block alleviates
the problem that the up-sampling through a single deconvolution
loses information. By taking global information into consideration,
the up-sampling block is able to recover more accurate details.

\section{Results and Discussion}

We perform experiments on the 3D multimodality isointense infant brain MR image segmentation task to evaluate our non-local U-Nets. The task is to perform automatic segmentation of MR images into cerebrospinal fluid~(CSF), gray matter~(GM) and white matter~(WM) regions. We first introduce the baseline model and the evaluation methods used in our experiments. Then the training and inference processes are described. We provide comparison results in terms of both effectiveness and efficiency, and conduct ablation studies to demonstrate that how each global aggregation block in our non-local U-Nets improves the performance. In addition, we explore the trade-off between the inference speed and accuracy based on different overlapping step sizes, and analyze the impact of patch size. The experimental code and dataset information have been made publicly available~\footnote{\url{https://github.com/divelab/Non-local-U-Nets}}.

\subsection{Experimental Setup}

We use CC-3D-FCN~\cite{nie20183} as our baseline.
CC-3D-FCN is a 3D fully convolutional network~(3D-FCN) with
convolution and concatenate~(CC) skip connections, which is designed
for 3D multimodality isointense infant brain image segmentation. It
has been shown to outperform traditional machine learning methods,
such as FMRIB's automated segmentation
tool~(FAST)~\cite{zhang2001segmentation}, majority
voting~(MV), random forest~(RF)~\cite{criminisi2013decision} and random forest with
auto-context model~(LINKS)~\cite{wang2015links}. Moreover, studies
in~\cite{nie20183} has showed the superiority of CC-3D-FCN to
previous deep learning models, like 2D, 3D
CNNs~\cite{zhang2015deep}, DeepMedic~\cite{kamnitsas2017efficient},
and the original 3D U-Net~\cite{cciccek20163d}. Therefore, it is
appropriate to use CC-3D-FCN as the baseline of our experiments.
Note that our dataset is different from that in~\cite{nie20183}.

In our experiments, we employ the Dice ratio~(DR) and propose the 3D modified
Hausdorff distance~(3D-MHD) as the evaluation metrics. These two
methods evaluate the accuracy only for binary segmentation tasks, so
it is required to transform the 4-class segmentation map predicted by our model into
4 binary segmentation maps for evaluation. That is, a 3D binary
segmentation map should be constructed for each class, where 1 denotes the voxel
in the position belongs to the class and 0 means the opposite. In
our experiments, we derive binary segmentation maps directly from
4-class segmentation maps. The evaluation is performed on binary
segmentation maps for CSF, GM and WM.

Specifically, let $P$ and $L$ represent the predicted binary segmentation map for one class
and the corresponding ground truth label, respectively. The DR is given by
$DR=2|P \cap L|/(|P|+|L|)$,
where $|\cdot|$ denotes the number of 1's in a segmentation map and $|P \cap
L|$ means the number of 1's shared by $P$ and $L$. Apparently, DR is a value
in $[0,1]$ and a larger DR indicates a more accurate segmentation.

\begin{table*}[!ht]
	\centering
	\caption{Comparison of segmentation performance between our proposed model
		and the baseline model in terms of DR. The leave-one-subject-out
		cross-validation is used. Larger values indicate better performance.}
	\label{table:results_baseline_dr}
	\begin{tabular}{ l | c | c | c | c }
		\toprule
		\textbf{Model} & \textbf{CSF} & \textbf{GM} & \textbf{WM} & \textbf{Average} \\
		\midrule
		Baseline
		& 0.9250$\pm$0.0118
		& 0.9084$\pm$0.0056
		& 0.8926$\pm$0.0119
		& 0.9087$\pm$0.0066 \\
		Non-local U-Net
		& \textbf{0.9530$\pm$0.0074}
		& \textbf{0.9245$\pm$0.0049}
		& \textbf{0.9102$\pm$0.0101}
		& \textbf{0.9292$\pm$0.0050} \\
		\bottomrule
	\end{tabular}
\end{table*}

\begin{table*}[!ht]
	\centering
	\caption{Comparison of segmentation performance between our proposed model
		and the baseline model in terms of 3D-MHD. The leave-one-subject-out
		cross-validation is used. Smaller values indicate better performance. Note that 3D-MHD gives different results from MHD.}
	\label{table:results_baseline_mhd}
	\begin{tabular}{ l | c | c | c | c }
		\toprule
		\textbf{Model} & \textbf{CSF} & \textbf{GM} & \textbf{WM} & \textbf{Average} \\
		\midrule
		Baseline
		& 0.3417$\pm$0.0245
		& 0.6537$\pm$0.0483
		& 0.4817$\pm$0.0454
		& 0.4924$\pm$0.0345 \\
		Non-local U-Net
		& \textbf{0.2554$\pm$0.0207}
		& \textbf{0.5950$\pm$0.0428}
		& \textbf{0.4454$\pm$0.0040}
		& \textbf{0.4319$\pm$0.0313} \\
		\bottomrule
	\end{tabular}
\end{table*}

\begin{table*}[!t]
	\centering
	\caption{Comparison of segmentation performance on the 13 testing subjects of iSeg-2017 between our proposed model
		and the baseline model in terms of DR. Larger values indicate better performance.}
	\label{table:iseg_results}
	\begin{tabular}{ l | c | c | c }
		\toprule
		\textbf{Model} & \textbf{CSF} & \textbf{GM} & \textbf{WM} \\
		\midrule
		Baseline
		& 0.9324$\pm$0.0067
		& 0.9146$\pm$0.0074
		& 0.8974$\pm$0.0123 \\
		Non-local U-Net
		& \textbf{0.9557$\pm$0.0060}
		& \textbf{0.9219$\pm$0.0089}
		& \textbf{0.9044$\pm$0.0153} \\
		\bottomrule
	\end{tabular}
\end{table*}

The modified Hausdorff distance~(MHD)~\cite{dubuisson1994modified} is
designed to compute the similarity between two objects. Here, an object is a
set of points where a point is represented by a vector. Specifically, given
two sets of vectors $A$ and $B$, MHD is computed by
$MHD=\max(d(A,B),d(B,A))$,
where the distance between two sets is defined as
$d(A,B)=1/|A|\sum_{a \in A}{d(a,B)}$,
and the distance between a vector and a set is defined as
$d(a,B)=\min_{b \in B}||a-b||$.
Previous studies~\cite{wang2015links,zhang2015deep,nie20183}
applied MHD for evaluation by treating a 3D $D \times H \times W$
map as $H \times W$ $D$-dimensional vectors. However, there are two
more different ways to vectorize the 3D map, depending on the
direction of forming vectors, \emph{i.e.,} $D \times H$
$W$-dimensional vectors and $D \times W$ $H$-dimensional vectors.
Each vectorization leads to different evaluation results by MHD. To
make it a direction-independent evaluation metric as DR, we define
3D-MHD, which computes the averaged MHD based on the three different
vectorizations. A smaller 3D-MHD indicates a
higher segmentation accuracy.

\subsection{Training and Inference Strategies}

Our proposed non-local U-Nets apply Dropout~\cite{srivastava2014dropout} with
a rate of 0.5 in each global aggregation block and the output block
before the final $1 \times 1 \times 1$ convolution. A weight
decay~\cite{krogh1992simple} with a rate of $2e-6$ is also employed.
To train the model, we use randomly cropped small patches. In this
way, we obtain sufficient training data and the requirement on
memory is reduced. No extra data augmentation is needed. The
experimental results below suggest that patches
with a size of $32^3$ leads to the best
performance. The batch size is set to 5. The Adam
optimizer~\cite{kingma2014adam} with a learning rate of 0.001 is
employed to perform the gradient descent algorithm.

In the inference process, following~\cite{nie20183}, we extract
patches with the same size as that used in training. For example, to
generate $32^3$ patches for inference, we slide a
window of size $32^3$ through the original image
with a constant overlapping step size. The overlapping step size
must be smaller than or equal to the patch size, in order to
guarantee that extracted patches cover the whole image.
Consequently, prediction for all these patches provides segmentation
probability results for every voxel in the original image. For
voxels that receive multiple results due to overlapping, we average
them to produce the final prediction. The overlapping step size is
an important hyper-parameter affecting the inference speed and the
segmentation accuracy. A smaller overlapping step size results in
better accuracy, but increases the inference time as more patches
are generated. We explore the trade-off in our experiments.

\subsection{Comparison with the Baseline}\label{sec:baseline}

We compare our non-local U-Nets with the baseline on our dataset.
Following~\cite{nie20183}, the patch size is set to $32^3$
and the overlapping step size for inference is set to $8$. To remove the
bias of different subjects, the leave-one-subject-out cross-validation is
used for evaluating segmentation performance. That is, for 10 subjects in our
dataset, we train and evaluate models 10 times correspondingly. Each time one
of the 10 subjects is left out for validation and the other 9 subjects are
used for training. The mean and standard deviation of segmentation
performance of the 10 runs are reported.

\begin{figure}[!t]
	\centering
	\includegraphics[width=0.95\columnwidth]{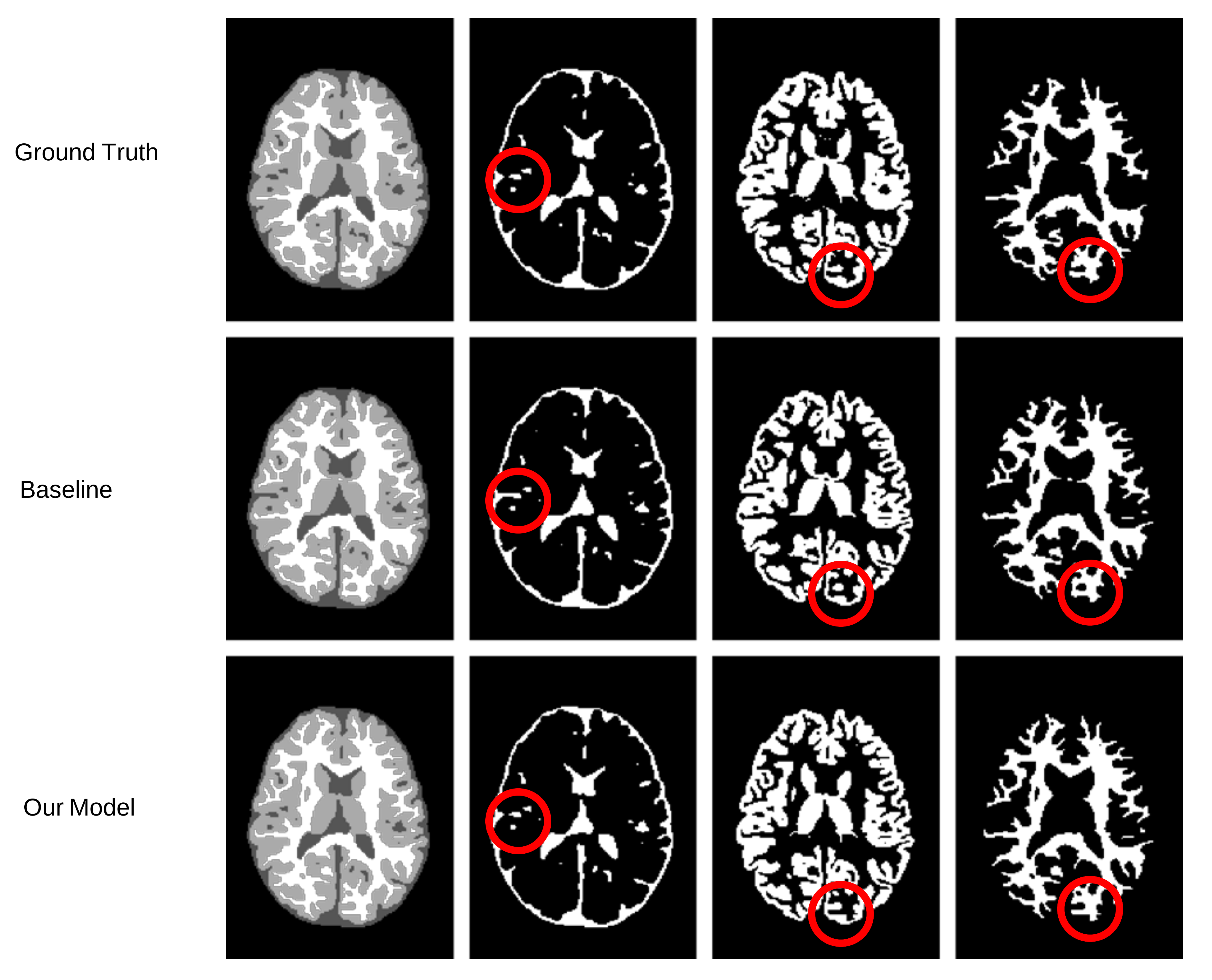}
	\caption{Visualization of the segmentation results on the $10^{th}$ subject
		by our proposed model and the baseline model. Both models are trained on the
		first 9 subjects. The first column shows the original segmentation maps. The
		second, third and fourth columns show the binary segmentation maps for CSF,
		GM and WM, respectively.}
	\label{fig:results_visual}
\end{figure}

\begin{figure}[!t]
	\centering
	\includegraphics[width=0.75\columnwidth]{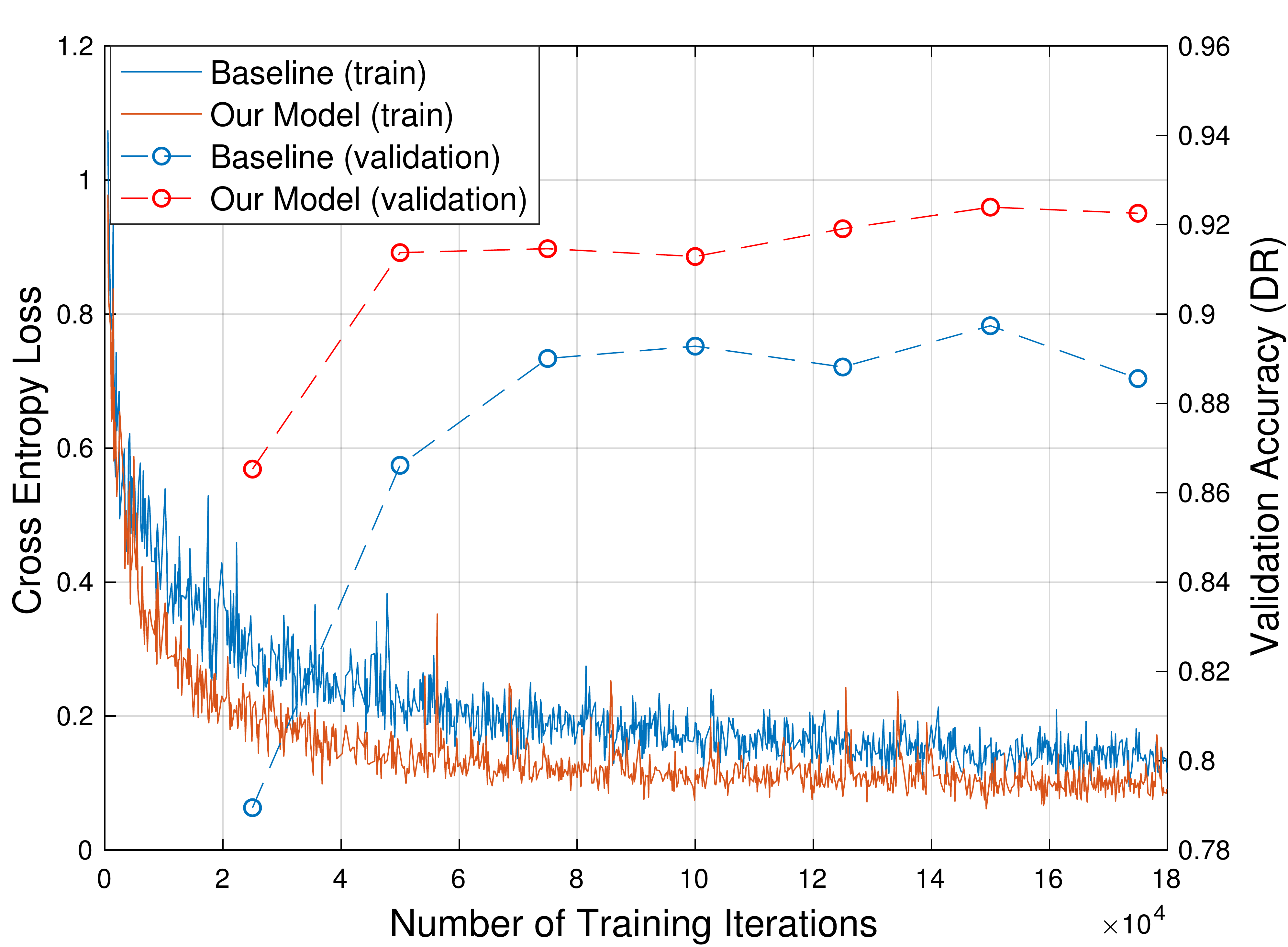}
	\caption{Comparison of training processes and validation results between our proposed model and the
		baseline model when training on the first 9 subjects and using the $10^{th}$ subject for validation.}
	\label{fig:results_training}
\end{figure}

Tables~\ref{table:results_baseline_dr}
and~\ref{table:results_baseline_mhd} provide the experimental
results. In terms of both evaluation metrics, our non-local U-Nets achieve
significant improvements over the baseline model. Due to the small
variances of the results, we focus on one of the 10 runs for
visualization and ablation studies, where the models are trained on the
first 9 subjects and evaluated on the $10^{th}$ subject. A
visualization of the segmentation results in this run is given by
Fig.~\ref{fig:results_visual}. By comparing the areas in red
circles, we can see that our model is capable of catching more
details than the baseline model. We also visualize the training
processes to illustrate the superiority of our model.
Fig.~\ref{fig:results_training} shows the training and validation
curves in this run of our model and the baseline model,
respectively. Clearly, our model converges faster to a lower
training loss. In addition, according to the better validation
results, our model does not suffer from over-fitting.

To further show the efficiency of our proposed model, we compare the
number of parameters as reported in Table~\ref{table:num_params}.
Our model reduces $28\%$ parameters compared to CC-3D-FCN and
achieves better performance. A comparison of inference time is also
provided in Table~\ref{table:infer_time}. The settings of our device
are - GPU: Nvidia Titan Xp 12GB; CPU: Intel Xeon E5-2620v4 2.10GHz;
OS: Ubuntu 16.04.3 LTS.

Since our data has been used as the training data in the iSeg-2017
challenge, we also compare the
results evaluated on the 13 testing subjects in
Table~\ref{table:iseg_results}. According to the leader board, our model
achieves one of the top performances. Results in terms of DR are reported since
it is the only shared evaluation metric.

\begin{table}[!t]
	\centering
	\caption{Comparison of the number of parameters between our proposed model
		and the baseline model.}
	\label{table:num_params}
	\begin{tabular}{ l | c }
		\toprule
		\textbf{Model} & Number of Parameters \\
		\midrule
		Baseline  & 2,534,276 \\
		Non-local U-Net & \textbf{1,821,124} \\
		\bottomrule
	\end{tabular}
\end{table}

\begin{table}[!t]
	\centering
	\caption{Comparison of inference time between our proposed model and the baseline model. The leave-one-subject-out cross-validation is used. The patch size is set to $32^3$ and the overlapping step size for inference is set to $8$.}
	\label{table:infer_time}
	\begin{tabular}{ l | c }
		\toprule
		\textbf{Model} & Inference Time (min) \\
		\midrule
		Baseline  & 3.85$\pm$0.15 \\
		Non-local U-Net  & \textbf{3.06$\pm$0.12} \\
		\bottomrule
	\end{tabular}
\end{table}

\begin{table}[!t]
	\centering
	\caption{Ablation study by comparing segmentation performance between
		different models in terms of DR. All models are trained on the first 9
		subjects and evaluated on the $10^{th}$ subject. Larger values indicate
		better performance. Details of models are provided in the text.}
	\label{table:results_ablation_dr}
	\resizebox{.95\columnwidth}{!}{
	\begin{tabular}{ l | c | c | c | c }
		\toprule
		\textbf{Model} & \textbf{CSF} & \textbf{GM} & \textbf{WM} & \textbf{Average} \\
		\midrule
		Model1       & \textbf{0.9585} & 0.9099 & 0.8625 & 0.9103 \\
		Model2       & 0.9568 & 0.9172 & 0.8728 & 0.9156 \\
		Model3       & 0.9576 & 0.9198 & 0.8749 & 0.9174 \\
		Model4       & 0.9578 & 0.9210 & 0.8769 & 0.9186 \\
		Model5       & 0.9554 & 0.9225 & 0.8804 & 0.9194 \\
		Non-local U-Net    & 0.9572 & \textbf{0.9278} & \textbf{0.8867} & \textbf{0.9239} \\
		\bottomrule
	\end{tabular}}
\end{table}

\begin{table}[!t]
	\centering
	\caption{Ablation study by comparing segmentation performance between
		different models in terms of 3D-MHD. All models are trained on the first 9
		subjects and evaluated on the $10^{th}$ subject. Smaller values indicate
		better performance. Note that 3D-MHD gives different results from MHD. Details of models are provided in
		the text.}
	\label{table:results_ablation_mhd}
	\resizebox{.95\columnwidth}{!}{
	\begin{tabular}{ l | c | c | c | c }
		\toprule
		\textbf{Model} & \textbf{CSF} & \textbf{GM} & \textbf{WM} & \textbf{Average} \\
		\midrule
		Model1       & \textbf{0.2363} & 0.6277 & 0.4705 & 0.4448 \\
		Model2       & 0.2404 & 0.6052 & 0.4480 & 0.4312 \\
		Model3       & 0.2392 & 0.5993 & 0.4429 & 0.4271 \\
		Model4       & 0.2397 & 0.5926 & 0.4336 & 0.4220 \\
		Model5       & 0.2444 & 0.5901 & 0.4288 & 0.4211 \\
		Non-local U-Net    & 0.2477 & \textbf{0.5692} & \textbf{0.4062} & \textbf{0.4077} \\
		\bottomrule
	\end{tabular}}
\end{table}

\subsection{Ablation Studies of Different Modules}\label{sec:ablation}

We perform ablation studies to show the effectiveness of each part of our
non-local U-Nets. Specifically, we compare the following models:

\textbf{Model1} is a 3D U-Net without short-range residual connections.
Down-sampling and up-sampling are implemented by convolutions and
deconvolutions with a stride of 2, respectively. The bottom block is simply a
convolutional layer. Note that the baseline model, CC-3D-FCN, has showed improved performance over 3D U-Net~\cite{nie20183}. However, the original 3D U-Net was not designed for this task~\cite{cciccek20163d}. In our experiments, we appropriately set the hyperparameters of 3D U-Net and achieve better performance.

\textbf{Model2} is Model1 with short-range residual connections, \emph{i.e.},
the blocks in Fig.~\ref{fig:residual}(a) and (b) are applied. The bottom
block and up-sampling blocks are the same as those in Model1.

\textbf{Model3} replaces the first up-sampling block in Model2 with the block
in Fig.~\ref{fig:residual}(d).

\textbf{Model4} replaces both up-sampling blocks in Model2 with the block in
Fig.~\ref{fig:residual}(d).

\textbf{Model5} replaces the bottom block in Model2 with the block in
Fig.~\ref{fig:residual}(c).

All models are trained on the first 9 subjects. We report the segmentation
performance on the $10^{th}$ subject in Table~\ref{table:results_ablation_dr}
and Table~\ref{table:results_ablation_mhd}. The results demonstrate how different
global aggregation blocks in our non-local U-Nets improve the performance.

\subsection{Impact of the Overlapping Step Size}\label{sec:overlap}

\begin{figure}[!t]
	\centering
	\includegraphics[width=0.7\columnwidth]{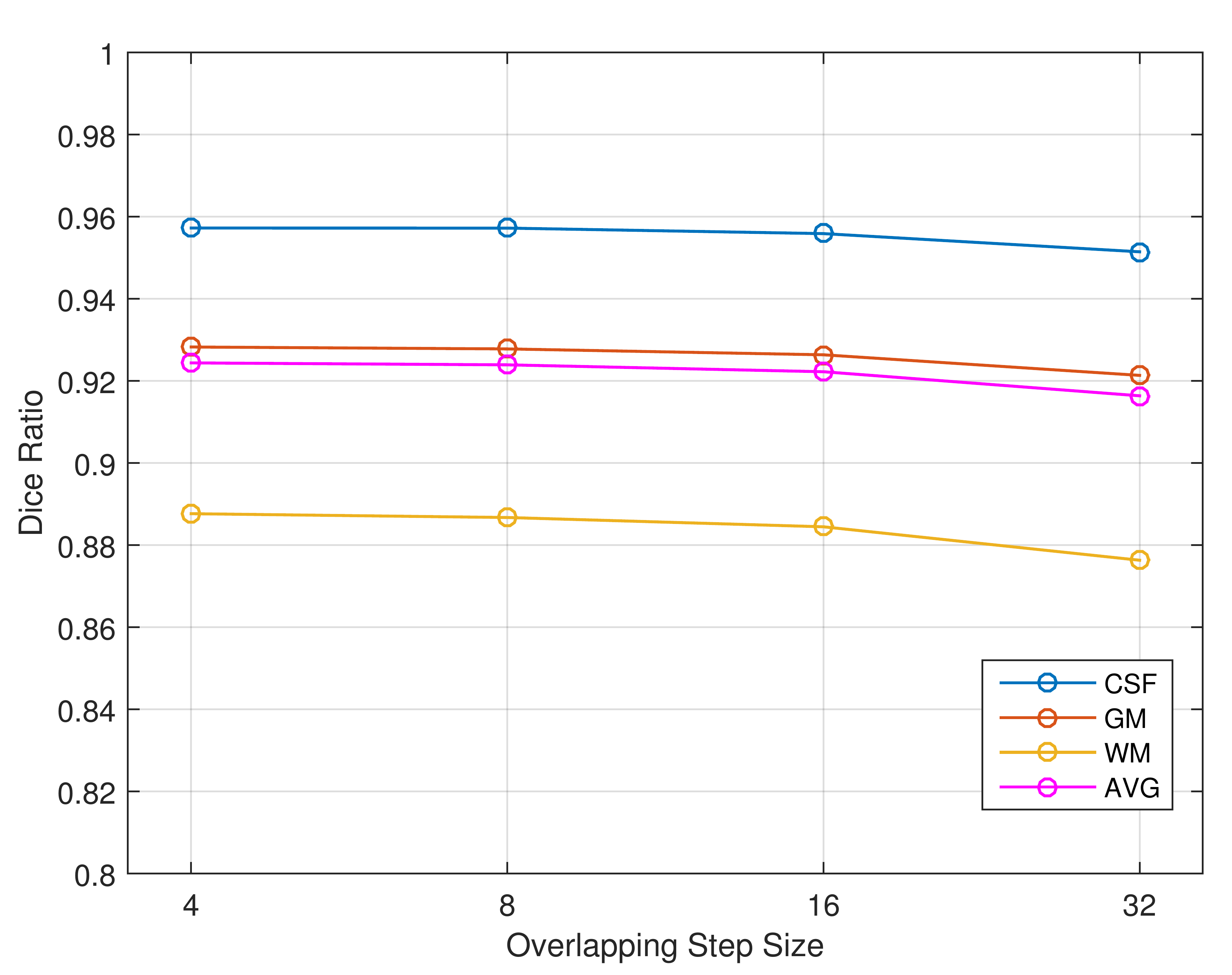}
	\caption{Changes of segmentation performance in terms of DR, with respect to
		different overlapping step sizes during inference. The model is trained on the
		first 9 subjects and evaluated on the $10^{th}$ subject.}
	\label{fig:results_overlap_dr}
\end{figure}

\begin{figure}[!t]
	\centering
	\includegraphics[width=0.75\columnwidth]{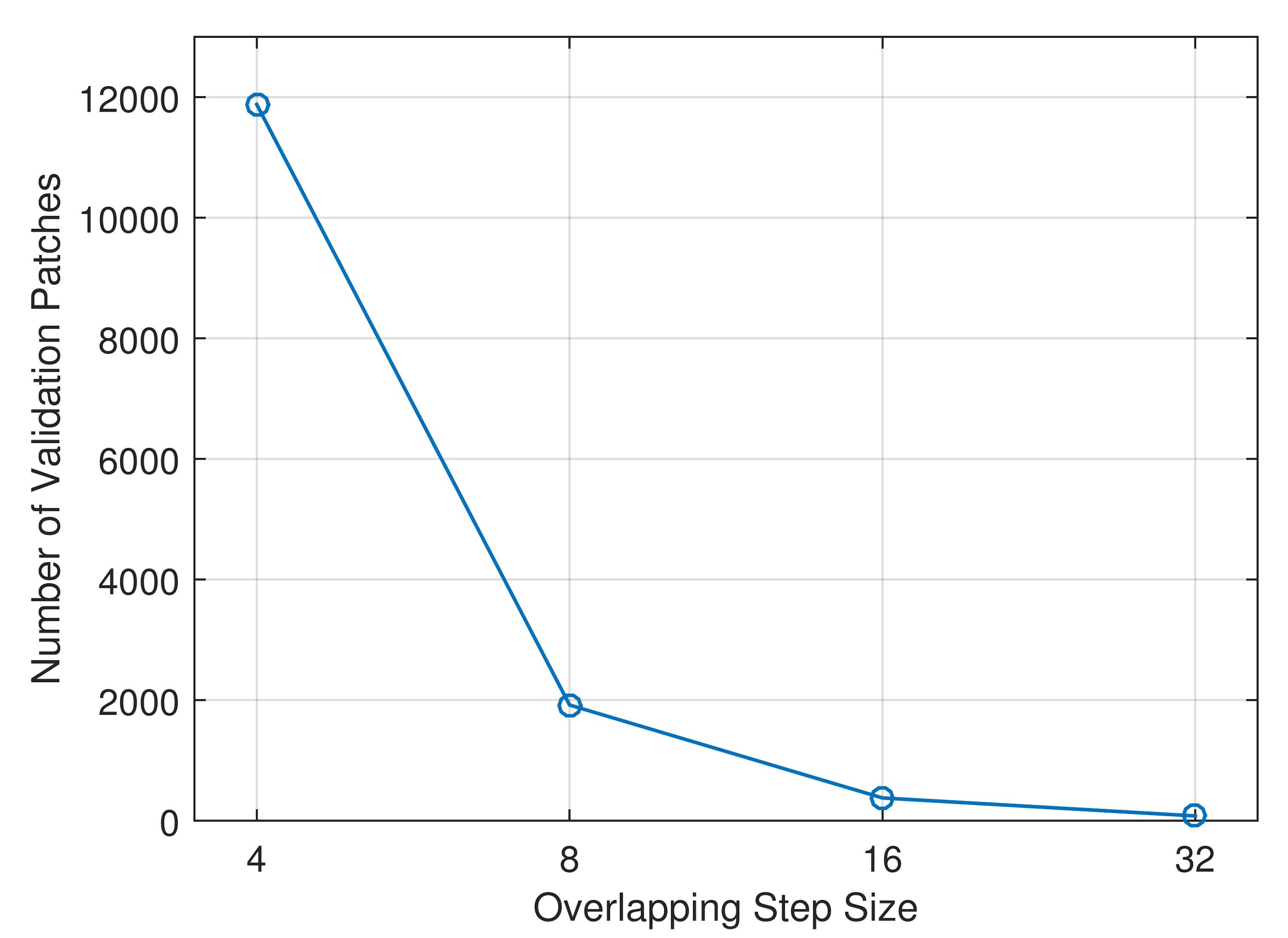}
	\caption{Changes of the number of validation patches for the $10^{th}$
		subject, with respect to different overlapping step sizes during inference.}
	\label{fig:results_overlap_time}
\end{figure}

As discussed above, a small overlapping step size
usually results in better segmentation, due to the ensemble effect.
However, with a small overlapping step size, the model has to perform
inference for more validation patches and thus decreases the inference speed. We
explore the trade-off in our non-local U-Nets by setting the overlapping step sizes to 4,
8, 16, 32, respectively. Again, we train our model on the first 9 subjects and
perform evaluation on the $10^{th}$ subject. The patch size is set to $32^3$.
According to the overlapping step sizes, 11880, 1920,
387, 80 patches need to be processed during inference, as shown in
Fig.~\ref{fig:results_overlap_time}. In addition, Fig.~\ref{fig:results_overlap_dr}
plots the changes of segmentation performance in terms of DR. Obviously, 8 and
16 are good choices that achieve accurate and fast segmentation results.

\subsection{Impact of the Patch Size}\label{sec:patch}

\begin{figure}[!t]
	\centering
	\includegraphics[width=0.75\columnwidth]{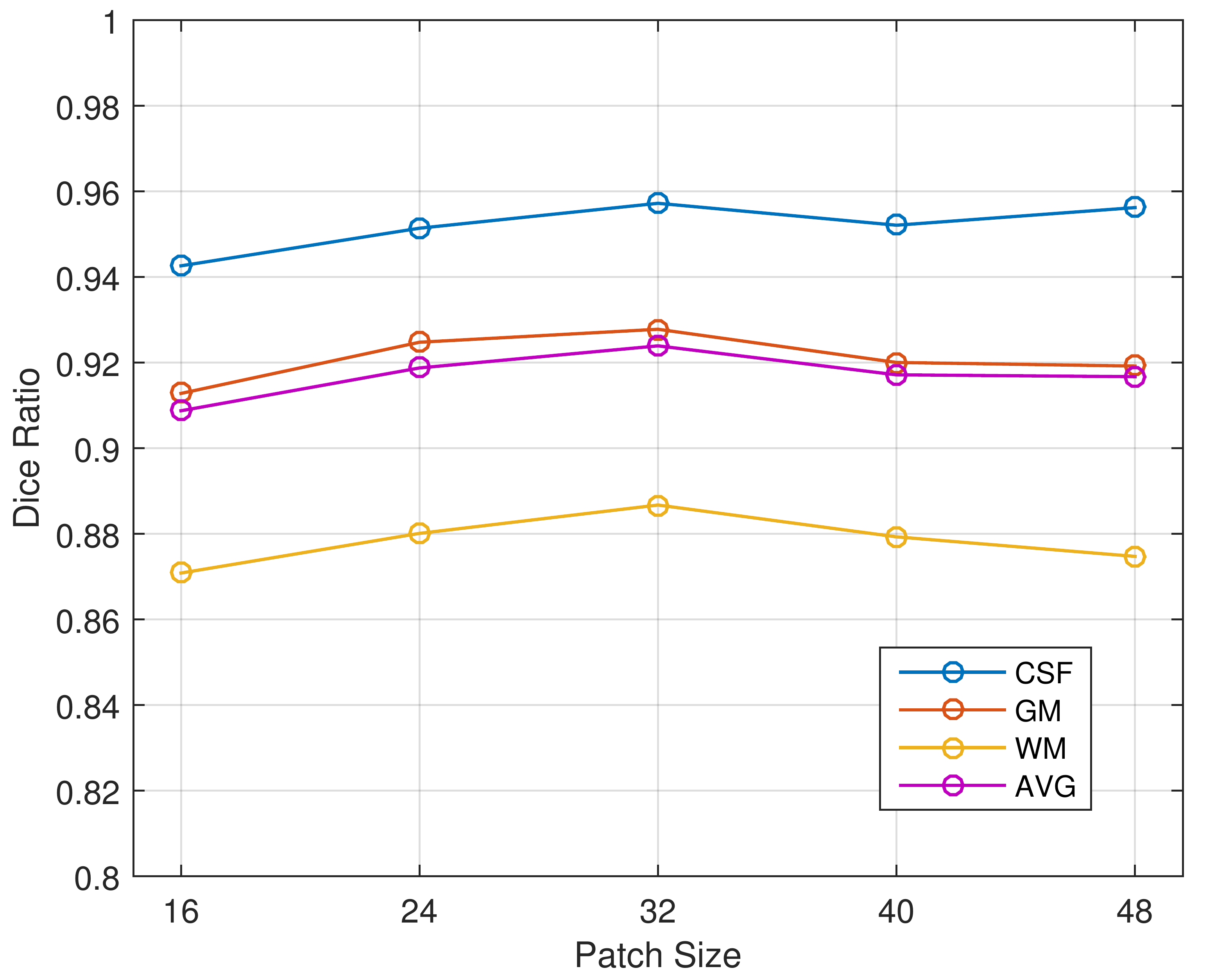}
	\caption{Changes of segmentation performance in terms of DR, with respect to
		different patch sizes. The model is trained on the first 9 subjects and
		evaluated on the $10^{th}$ subject.}
	\label{fig:results_patch_dr}
\end{figure}

The patch size affects the total number of distinct training samples.
Meanwhile, it controls the range of available global information when
performing segmentation for a patch. To choose the appropriate patch
size for the non-local U-Nets, we perform a grid search by training on the first 9
subjects and evaluating on the $10^{th}$ subject with the overlapping step
size of 8. Experiments are conducted with five different patch sizes:
$16^3$, $24^3$, $32^3$, $40^3$, $48^3$. The results are provided in
Fig.~\ref{fig:results_patch_dr}, where $32^3$ obtains the best
performance and is selected as the default setting of our model.

\section{Conclusion}

In this work, we propose the non-local U-Nets for biomedical image segmentation. As pointed out, prior U-Net based models do not have an efficient and effective way to aggregate global information by using stacked local operators only, which limits their performance. To address these problems, we propose a global aggregation block which can be flexibly used in U-Net for size-preserving, down-sampling and up-sampling processes. Experiments on the 3D multimodality isointense infant brain MR image segmentation task show that, with global aggregation blocks, our non-local U-Nets outperform previous models significantly with fewer parameters and faster computation.

\section{ Acknowledgments}

This work was supported by National Science Foundation grants IIS-1908166 and IIS-1900990, and Defense Advanced Research Projects Agency grant N66001-17-2-4031.

\bibliographystyle{aaai}
\bibliography{reference}

\end{document}